\documentclass{article}

\usepackage[final]{IIBPSL_2020}

\usepackage[utf8]{inputenc} % allow utf-8 input
\usepackage[T1]{fontenc}    % use 8-bit T1 fonts
\usepackage{hyperref}       % hyperlinks
\usepackage{url}            % simple URL typesetting
\usepackage{booktabs}       % professional-quality tables
\usepackage{amsfonts}       % blackboard math symbols
\usepackage{nicefrac}       % compact symbols for 1/2, etc.
\usepackage{microtype}      % microtypography
\usepackage{graphicx}
\usepackage{color}
\graphicspath{ {./images/} }
\usepackage{adjustbox}

\title{Solving Physics Puzzles by Reasoning about Paths}

\author{%
  Augustin Harter\\
  Bielefeld University\\
  \texttt{aharter@techfak.uni-bielefeld.de}\\
  \And
  Andrew Melnik \\
  Bielefeld University\\
  \texttt{andrew.melnik.papers@gmail.com}\\
  \And
  \hspace*{17mm}Gaurav Kumar\\
  \hspace*{17mm}Bielefeld University\\
  \And
  \hspace*{15mm}Dhruv Agarwal\\
  \hspace*{15mm}Indian Institute of Information Technology\\
  \And
  \hspace*{-17mm}Animesh Garg\\
  \hspace*{-17mm}University of Toronto, Vector Institute, Nvidia\\
  \And
  \hspace*{-5mm}Helge Ritter\\
  \hspace*{-5mm}Bielefeld University\\
  \\
}

\begin{document}

\maketitle

\vspace*{-4mm}\begin{abstract} 
We propose a new deep learning model for goal-driven tasks that require intuitive physical reasoning and intervention in the scene to achieve a desired end goal. Its modular structure is motivated by hypothesizing a sequence of intuitive steps that humans apply when trying to solve such a task. The model first predicts the path the target object would follow without intervention and the path the target object should follow in order to solve the task. Next, it predicts the desired path of the action object and generates the placement of the action object. All components of the model are trained jointly in a supervised way; each component receives its own learning signal but learning signals are also backpropagated through the entire architecture. To evaluate the model we use PHYRE - a benchmark test for goal-driven physical reasoning in 2D mechanics puzzles.
\end{abstract}

\begin{figure}[h]
  \centering
  \includegraphics[scale=0.24]{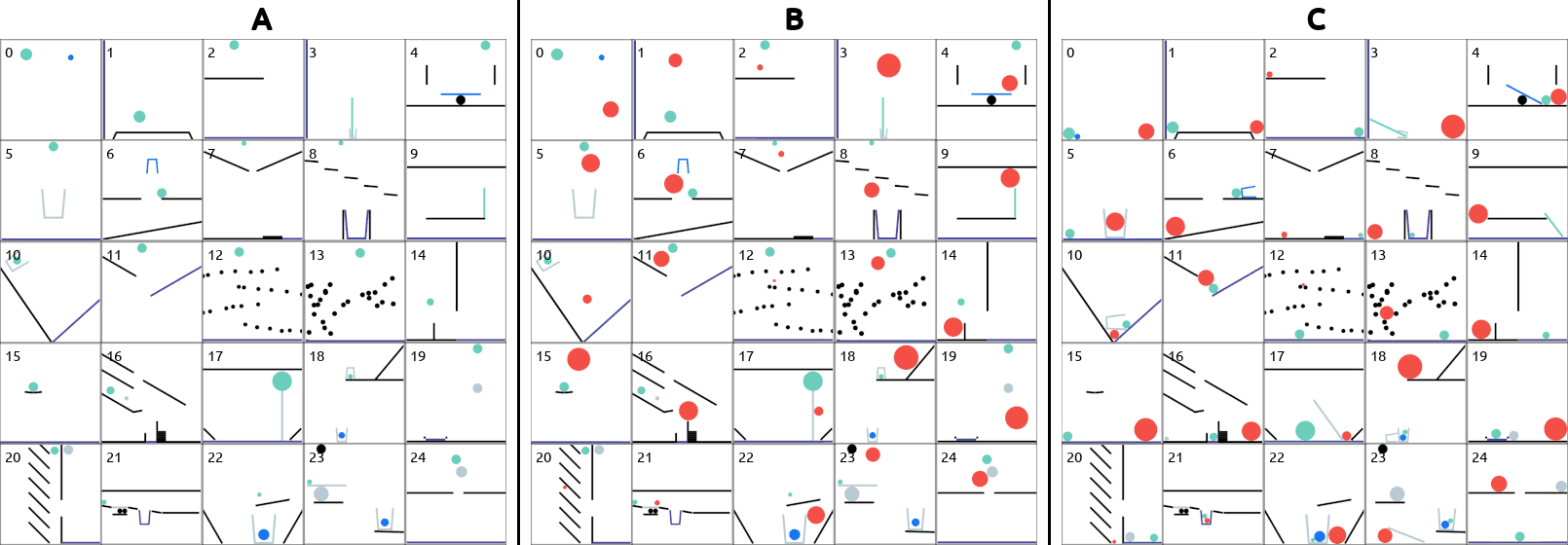}
  \caption{25 templates of PHYRE-B (BALL) puzzles \cite{bakhtin2019phyre} \textit{A:} initial puzzle scenes. Task: a green target-object has to touch a blue goal-object for 3 seconds. \textit{B:} placing red action-balls that solve the puzzles. \textit{C:} solved puzzle scenes (green target-objects are in firm contact with blue goal-objects). Human players can try to solve these PHYRE tasks here: (https://player.phyre.ai).}
  \label{fig-templates}
\end{figure}

\section{Introduction}

Consider the following physics puzzle: select a location for dropping a ball so that it hits a number of other objects in such a manner that a "target" object gets firm contact with a "goal" object (Fig. \ref{fig-templates}). PHYRE benchmark\footnotemark \cite{bakhtin2019phyre} offers a large set of physics challenges in the described format. The ability of humans to solve even intricate instances of such puzzles reflects a high level of physics cognition that so far is unparalleled in machines.

\footnotetext{https://player.phyre.ai}

\section{Methods}

\subsection{Action generation model}
\label{model}

\begin{figure}[h]
  \centering
  \vspace{-5mm}\hspace*{-6mm}\includegraphics[scale=0.69]{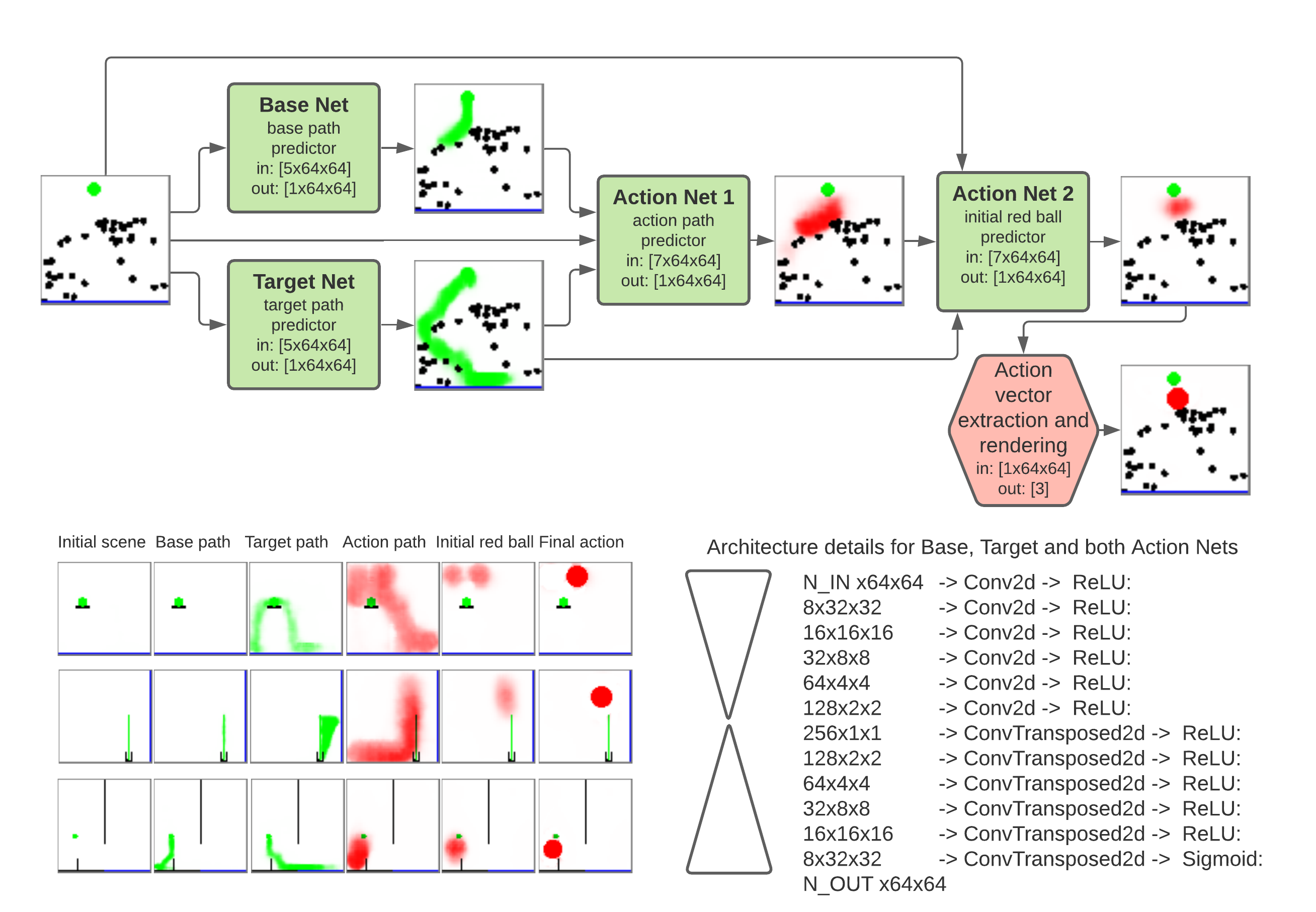}
  \caption{  \textit{Top:} Action generation pipeline. NNs modules are highlighted with green rectangles. The task's initial scene is presented to the agent as five bitmap channels; one channel for each object class: Green target-object, blue dynamic goal-object, blue static goal-object, dynamic grey objects, static black objects.  \textit{Bottom left:} Model prediction examples. All examples of the generated \textit{final action} in the figure solve the corresponding tasks.  \textit{Bottom right:} Model architecture details: Every Conv2d and ConvTransposed2d Layer has a kernel size of 4x4, stride of 2 and padding of 1.}
  \label{fig-pipe-action}
\end{figure}

Here we describe the building blocks of our NN model\footnotemark outlined in Figure~\ref{fig-pipe-action}, and compare them to the introspection of players trying to solve such a puzzle:

\footnotetext{https://github.com/ndrwmlnk/PHYRE-Reasoning-about-Paths}

\begin{enumerate}
\item \textit{Base Net}. Humans can get useful information by imagining how the scene will develop without the solving intervention. Therefore, we train a \textit{Base Net} neural network (NN) using the PHYRE simulator to predict the path (\textit{Base Path}) of the green target-object (Fig. \ref{fig-templates}A) when there is no interaction with the red action-ball. The \textit{Base Net} input consists of the five channel bitmap of the initial scene. Target output is the probability density of 2D-points predicted to be traversed by the \textit{Base Path}. The training procedure is described in detail in the section \ref{sec:training} \textit{Training}.

\item \textit{Target Net}. Human players can imagine the path the green target-object should follow in order to solve the puzzle on a set of acquired intuitions: Getting closer to the goal is better as well as following the direction of gravity and obeying interaction constraints. Therefore, we train a \textit{Target Net} (Figure \ref{fig-pipe-action}) to generate possible solution paths (\textit{Target Path}) of the green target-object without information about the red action-ball. This gives the \textit{Target Net} the freedom to "dream up" \textit{Target Paths} that appear "interaction incomplete" since they become physically valid only through some, so far unknown, interaction with the red action-ball. Input to the \textit{Target Net} is the initial scene 5-channel bitmap. Target output now is the probability map (2D-density of traversed points) of the \textit{Target Path}.

\item \textit{Action Net 1}. Humans are able to reason and imagine which trajectories the red action-ball has to take to "add" the missing interaction to turn a \textit{Base Path} into a potential \textit{Target Path}. Here humans might heuristically start from the point where \textit{Base Path} and \textit{Target Path} diverge. \textit{Action Net 1} generates possible \textit{Action-Ball Paths}. NN input: Initial scene 5-channel bitmap and 2-channel probability maps of the \textit{Base Path} from step 1 and the \textit{Target Path} from step 2. NN output: Probability map of the \textit{Action-Ball Path}.

\item \textit{Action Net 2} generates a probability map of the initial red action-ball position (Fig. \ref{fig-templates}B). NN input: Initial scene 5-channel bitmap and probability maps of the \textit{Target Path} from step 2 and the \textit{Action Path} from step 3.

\item Convert the red action-ball probability maps from step 4 to a 3-dim action vector, comprised of x, y and radius values. This is done with a non-learning algorithm, which randomly selects initial radius and initial position proposals (x, y) from pixels that are over a certain threshold value. Then it iteratively tries improve the overlap of the red action-ball rendered from a new action vector, and the probability map by selecting position and radius values within a close neighborhood. We sample 5 different initial positions and radius value and update each of them 5 times. We take these as the first 5 actions for solving the task. If all fail, we randomly sample from them and add Gaussian noise until the task is solved or the limit of 100 tries is reached.

\end{enumerate}

\subsection{Encoder-Decoder Hourglass-like model}
The artificial agent needs knowledge about the whole scene for reasoning about paths: Object interactions might be understood with local information but to propose a path which connects the green target-object and blue goal-object the whole scene needs to be considered. This motivated the following architecture for \textit{Base Net}, \textit{Target Net}, \textit{Action Net 1}, and \textit{Action Net 2}, illustrated in Figure \ref{fig-pipe-action}. A stack of convolutional layers 'folds' the input channels into a 256-dimensional encoding of the complete scene. Then a similar stack of transposed convolutional layers 'unfolds' this global encoding into the desired number of output channels, each having the same dimensionality as the input channels. We use \textit{ReLU} as the activation function for hidden layers and the \textit{sigmoid} function for the output layer to allow smooth prediction.

\subsection{Training}
\label{sec:training}
All 4 NNs of the model are trained jointly in a supervised way; each NN receives its own learning signal but learning signals are also backpropagated through the entire architecture. We collect \textit{BasePath}, \textit{TargetPath}, \textit{ActionPath} and the initial red action-ball bitmap channels from rollouts in the PHYRE simulator and use them to impose a cross-entropy loss between every pixel of the NNs output and the corresponding ground truth bitmaps. \textit{Action Net 1} and \textit{Action Net 2} receive predicted \textit{BasePath} and \textit{TargetPath} channels from \textit{BaseNet} and \textit{TargetNet} NNs. For training we use a data set that containes 10 solving rollouts per task, 80 tasks per template, and 25 PHYRE-B templates. We randomly shuffle the samples and split them into 625 batches with a batch size of 32 and train the NNs for 10 epochs.

\section{Results}

\subsection{Task Solving Performance}

\textit{PHYRE} uses the \textit{auccess} metric to score performance:
Agents can try up to 100 attempts per task and the area under the logarithmically scaled \textit{percentage of tasks solved} curve is the \textit{auccess} value. See \cite{bakhtin2019phyre} for more details. As described in \ref{model}, we generate 5 action proposals for each task based on the action ball prediction from \textit{Action Net 2}. If the proposed actions led to unsuccessful attempts, we sample further action vectors from a multivariate normal distribution centered at the original action proposals. We slowly increase its standard deviation during the 100 tasks (starting from 0.02), leading to increased deviation from the original action for later attempts. There are 25 templates in the PHYRE-B (BALL) problem \cite{bakhtin2019phyre}. In the within-template setting the model is evaluated on tasks that share the same template with training tasks. Such tasks differ in the placement and size of scene objects, but not in their otherwise structure. In the cross-template setting, the model is evaluated on tasks from templates that were not shown during training. Bakhtin et al. \cite{bakhtin2019phyre} use a "process that deterministically splits the tasks into 10 folds containing a training, validation, and test set" allowing fair comparison between agents and studies. Table \ref{indi-auccess} shows auccess values of our trained NN model which are collected individually for each template in each fold and then averaged over all 10 folds.

\begin{table}[h]
    \caption{Auccess (\textit{auc.}) and percentage (\textit{perc.}) of solved tasks after 10 attempts in PHYRE-B (BALL) templates (Fig. \ref{fig-templates})}
    \label{indi-auccess}
    \begin{adjustbox}{width=1\textwidth}
    \centering
    \begin{tabular}{l|ccccccccccccccccccccccccc|c}
    \toprule
    template&0&1&2&3&4&5&6&7&8&9&10&11&12&13&14&15&16&17&18&19&20&21&22&23&24& \textbf{mean} \\
    \midrule
    within \textit{auc.} & 80 & 89 & 87 & 95 & 79 & 82 & 86 & 62 & 69 & 77 & 31 & 53 & 52 & 65 & 84 & 78 & 39 & 41 & 69 & 45 & 46 & 29 & 30 & 45 & 42 & \textbf{62}\\
    within \textit{perc.} &86 &98 &90 &100   &86 &90&94 &62 &72&81 &27&57&55&70  &93 &83 &37&44 &82&51&48     &23 &27 &46&44 & \textbf{66}\\
    cross \textit{auc.} &51 &54 &67 &67 &57 &63& -&30  &24 &14 &14 &38  &48 &51 &15  &25 &14 &5 &37 &17 &24 &3&9 &6 &6 & \textbf{31}\\
    cross \textit{perc.} &50&51 &72 &70  &58 &67 & - &31 &22 &13 &11 &34 &50 &53 &11 &21 &12 &6 &35 &13&22 &0 &7 &4 &6 & \textbf{31}
    % \bottomrule
\end{tabular}%
\end{adjustbox}
\end{table}

% \footnote{https://player.phyre.ai}
% \footnotemark
% \footnotetext{https://player.phyre.ai}

\section{Discussion and related works}

Our action generation model (Table \ref{indi-auccess}) falls behind the baseline DQN's performance \cite{bakhtin2019phyre} of 77 within template \textit{auccess}, 81 within template \textit{percentage} of solved tasks after 10 attempts, 37 cross template \textit{auccess}, and 35 cross template \textit{percentage} of solved tasks after 10 attempts. But one important difference is to be noted: The baseline DQN \textit{only} uses a deep reinforcement learning setup to guess the success value of a batch of 10000 possible combinations of action parameters (x, y, radius) with a certain level of grid discretization, while our model performs a basic reasoning about developments in the scene and interactions about objects, thus capable of generating an informed action proposal in an explainable fashion. Our model can also be thought of as a kind of a jointly trained autoencoder for multi-modal fusion \cite{korthals2019jointly}.

Forward prediction for physical reasoning \cite{girdhar2020forward}, an error-based dynamic model learning \cite{bach2020error}, or continues-action-space policy gradient algorithms \cite{bach2020learn} are possible ways to improve performance and generalize learning. Allen et al. \cite{allen2019tools} introduced the “Virtual Tools” game to measure the capacity of human beings for flexible, creative tool use. The study proposes that the flexibility of human physical problem solving rests on an ability to imagine the effects of hypothesized actions. Our deep learning architecture (Fig.~\ref{fig-pipe-action}) can serve as a mechanism to imagine the effects of such hypothesized actions. Kurutach et al. \cite{kurutach2018learning} demonstrated on a rope manipulation example that a plausible sequence of observations evolving from its current configuration to a desired goal state can be used as a reference trajectory for control. In contrast, the PHYRE type problems have an additional constraint of one control-action per episode, and therefore requires a different NN architecture.

Melnik et al. \cite{melnik2019modularization} described a set of functional modules for specific types of interaction primitives, which are common to a broad range of arcade ball-game environments. Results of this case study in different Atari ball-games suggest that human-level performance can be achieved by a learning agent within a human amount of game experience (10-15 minutes game time) when a proper decomposition of an environment or a task is provided. However, automatization of such decomposition remains a challenging problem.

\section{The broader impact statement}

Human beings start to understand and reason about the goal-driven interaction of physical objects in the environment in early childhood, and therefore this skill also contributes to learning of other skills, e.g. language, logic, etc. Solving this problem in simpler 2D environments is likely to be an important step in tackling the difficult general problem in real-world 3D environments. In this work, we developed such a goal-driven intuitive-physics-reasoning NN model with strong generalization properties mirroring those of humans.

\medskip
\newpage
\bibliographystyle{unsrt}
\bibliography{main}

\end{document}